\documentclass{article}
\usepackage{ijcai24}

\usepackage{times}
\usepackage{url}
\usepackage[hidelinks]{hyperref}
\usepackage[utf8]{inputenc}
\usepackage[small]{caption}
\usepackage{graphicx}
\usepackage{amsmath}
\usepackage{amsthm}
\usepackage{booktabs}
\usepackage{algorithm}
\usepackage{algorithmic}
\usepackage[switch]{lineno}
\usepackage{tabularx}
\usepackage{cleveref}

\title{Adversarial Detection with a Dynamically Stable System}

\author{
Xiaowei Long$^1$
\and
Jie Lin$^2$\and
Xiangyuan Yang$^3$\\
\affiliations
$^{1,2,3}$Xi'an Jiaotong University\\
\emails
lxw912963765@stu.xjtu.edu.cn,
jielin@mail.xjtu.edu.cn,
ouyang\_xy@stu.xjtu.edu.cn
}

\begin{document}

\maketitle

\begin{abstract}
    Adversarial detection is designed to identify and reject maliciously crafted adversarial examples(AEs) which are generated to disrupt the classification of target models.  
    Presently, various input transformation-based methods have been developed on adversarial example detection, which typically rely on empirical experience and lead to unreliability against new attacks. 
    To address this issue, we propose and conduct a \textbf{Dynamically Stable System (DSS)}, which can effectively detect the adversarial examples from normal examples according to the stability of input examples. 
    Particularly, in our paper, the generation of adversarial examples is considered as the perturbation process of a Lyapunov dynamic system, and we propose an example stability mechanism, in which a novel control term is added in adversarial example generation to ensure that the normal examples can achieve dynamic stability while the adversarial examples cannot achieve the stability. 
    Then, based on the proposed example stability mechanism, a Dynamically Stable System (DSS) is proposed, which can utilize the \textit{disruptio}n and \textit{restoration} actions to determine the stability of input examples and detect the adversarial examples through changes in the stability of the input examples. 
    In comparison with existing methods in three benchmark datasets(MNIST, CIFAR10, and CIFAR100), our evaluation results show that our proposed DSS can achieve ROC-AUC values of 99.83\%, 97.81\% and 94.47\%, surpassing the state-of-the-art(SOTA) values of 97.35\%, 91.10\% and 93.49\% in the other 7 methods.
    
\end{abstract}

\section{Introduction}
Adversarial examples(AEs) are intentionally perturbed examples containing artificial noises, inducing misclassification in deep neural networks(DNNs). This susceptibility of DNNs to AEs has raised extensive concerns \cite{FawziFF18} due to the diverse security applications of DNNs, such as face recognition \cite{face-recognition}, autonomous driving \cite{LiuWYVZ23}, and the medical domain \cite{uwimana2021out}, etc. Consequently, an advanced defense method is necessary to mitigate the risk from adversarial examples.

A widely accepted hypothesis \cite{adversarial_examples} suggests that the success of AEs is attributed to their capability to shift the data flow from high-probability regions to low-probability regions outside the normal training domain of classifiers.
Two types of defense methods: adversarial training and adversarial detection have been proposed based on the hypothesis of data flow patterns. 
Initially, the adversarial training \cite{FGSM} is introduced, in which researchers incorporate AEs into the training data to broaden the training data flow, thereby enhancing the robustness to AEs.
Subsequently, adversarial detection emerges, aiming to extract manifold features from inputs to differentiate between benign and malicious inputs and subsequently reject the latter.
For the reason that adversarial training leads to a notable decrease in classification accuracy \cite{TsiprasSETM19} and generalization ability \cite{LaidlawF19}, adversarial detection is widely adopted, as it does not weaken the original performance of models.
Currently, many detection methods have been proposed and are mainly divided into mathematical statistical and variational analytical categories.
Mathematical statistics methods, like Kernel Density and Bayesian Uncertainty (KDBU) \cite{KDBU} and Local Intrinsic Dimensionality (LID) \cite{LID}, calculate density information based on the intermediate layer distribution of input data.
Furthermore, \cite{MD} introduces the Mahalanobis distance to improve the assessment of high-dimensional data, while Joint statistical Testing across DNN Layers for Anomalies (JTLA) \cite{JTLA} incorporates a meta-learning system to combine several existing methods.
However, these mathematical-statistical methods, which are static and lacking considering dynamic characteristics of inputs over the input transformations, perform weakly in the generalization domain. 

Variational analytical methods calculate the variations of the outputs in the target model to differentiate AEs and NEs when the input is transformed by rotation, flip, shift, etc. 
Feature Squeezing (FS) \cite{FS} involves bit reduction and shifting inputs, while Lightweight Bayesian Refinement (LIBRE) \cite{LIBRE} utilizes Bayesian network structures to extract the uncertainty of inputs and Expected Perturbation Score (EPS) \cite{EPS} introduces a diffusion model to augment input diversities.
However, these methods only empirically demonstrate the difference of output variations on both AEs and NEs in the target model, lacking theoretical support, resulting in unreliability to new attack strategies such as Square \cite{Square}. 

\begin{figure}[htbp]
    \centerline{\includegraphics[width=0.5\textwidth]{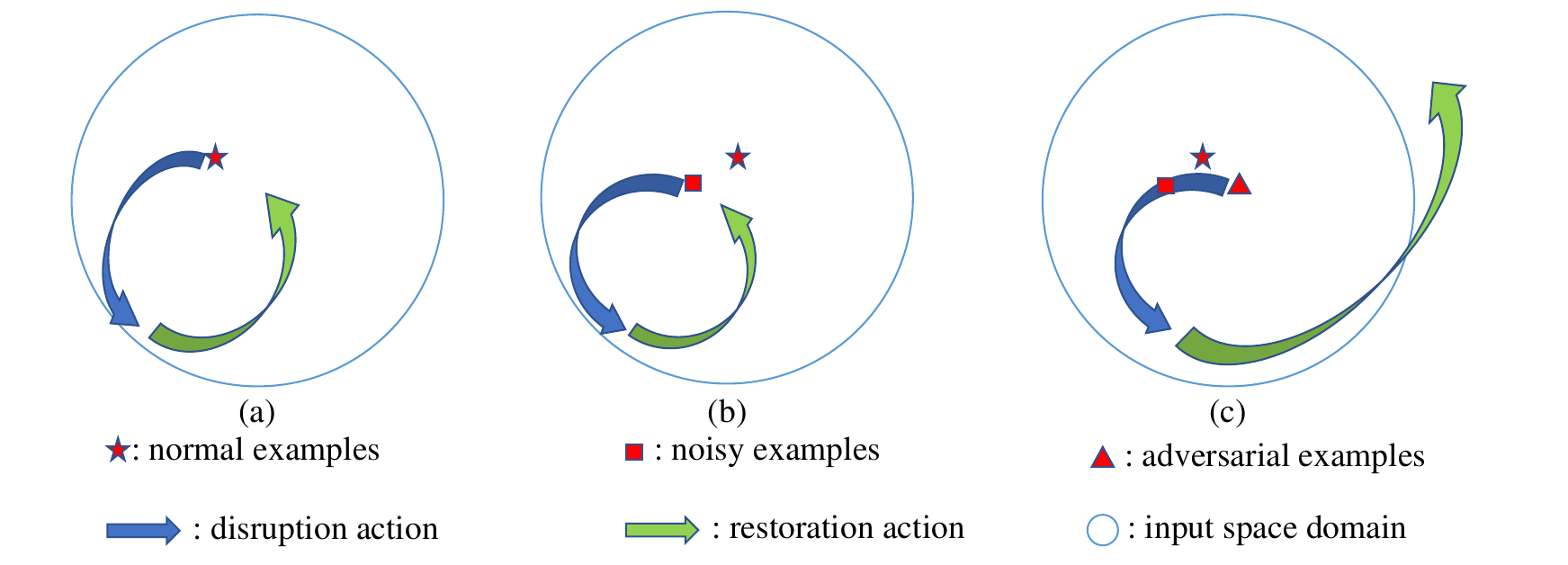}}
    \caption{The disruption and restoration actions through the stability module (a):the process of normal examples (b):the process of noisy examples (c):the process of adversarial examples}
    \label{fig1}
    \vspace{-5pt}
\end{figure}

To solve these issues, we construct the \textbf{Dynamically Stable System(DSS)} based on the Lyapunov stability theory.
Our system falls under the variational analytical category, as the stability module of the system involves disruption and restoration actions.
In the stability module, we iteratively disrupt and restore the inputs to obtain dynamic stability features.
Normal and noisy examples, when subjected to the disruption and restoration process, tend to maintain stability, as shown in Fig.\ref{fig1} (a) and (b).
On the contrary, AEs exhibit a tendency to diverge as the system progresses and will be detected based on this tendency, as shown in Fig.\ref{fig1} (c).
In the monitor module, we distinguish AEs from NEs based on different stability features provided by the stability module.

Our main contribution can be summarized as below:
\begin{itemize}
  \item Firstly, we prove that the process of adversarial examples generation can be considered as a Lyapunov dynamic stable system, and we propose an example stability mechanism, in which a novel control term is proposed to be involved in the generation. The proposed control term is proven to ensure the normal examples achieve the original examples stability. That is, with the control term, the stability of outputs of normal examples will be few changes to that of the original examples, while the stability of outputs of adversarial examples will achieve high alternations to that of the original examples.
  \item Then, based on the proposed example stability mechanism, we propose a \textbf{Dynamically Stable System}, which can effectively detect the adversarial examples according to the changes in the stability of our original examples. The proposed DSS consists of a stability module and a monitor module. In the stability module, the input original example is disturbed by the malicious gradient information with the proposed control term and then we introduce an inpainting model to repair the disturbed examples. In this way, the original normal examples will achieve similar stability to the repaired examples, while the adversarial examples will achieve much different stability to the repaired examples. Hence, the monitor module can detect the adversarial examples according to the changes in the stability of input original examples.
  \item Lastly, extensive experiments are conducted to demonstrate the effectiveness of our proposed DSS in three benchmark datasets(MNIST, CIFAR10, and CIFAR100). The evaluation results show that our DSS outperforms the best results, i.e., achieving the average AUC of 99.83\%, 97.81\% and 94.47\% respectively, in comparison with 7 existing adversarial detection methods. Furthermore, the evaluation in terms of generalization, intensity, ablation and sensitivity has been conducted as well and the results demonstrate the outstanding performance of our DSS compared to other methods.
\end{itemize}

\section{Related Work}
\subsection{Adversarial Attacks}
Adversarial attacks add artificial malicious perturbations invisible to human eyes on inputs, causing misclassifications of models.
\cite{FGSM} firstly proposed the FGSM method by loss gradient descents according to the linear ability assumption of neural networks. 
Based on the linear assumption, BIM \cite{BIM} utilized an iterative loop strategy with a smaller stepsize based on FGSM; 
PGD \cite{PGD} optimized the BIM by adding random initializations at each loop; 
APGD and AutoAttack \cite{APGD_AutoAttack} utilized the adaptive learning rate and a linear combination of several attacks, respectively.
Besides the gradient-based attacks, DeepFool \cite{DeepFool} calculated the classification boundaries to decide the attacking directions; 
CW \cite{CW} optimized the loss function and input space domains. 
Square \cite{Square} accomplished the black-box attack with thousands of queries. 
These attacks span various attack domains and are subsequently utilized to generate AEs.

\subsection{Adversarial Defenses}
Existing defense mechanisms primarily fall into two categories: adversarial training and adversarial detection. 
The prevailing strategy in adversarial training involves incorporating AEs into the model training dataset.
For instance, \cite{FGSM} initially introduced AEs into the training set, aiming to enhance the robustness of the model by learning the characteristics of AEs. 
Additionally, both \cite{PinotERCA20} and \cite{Dong00022} employed random normalization layer modules to improve the model robustness.
Meanwhile, \cite{LiWC0H22} conducted adversarial training with constrained gradients to avoid overfitting.
However, the model generalization ability is quite limited due to the addition of AEs into the training set. 

In the detection domain, \cite{KDBU} firstly utilized intermediate layer embeddings to construct probability density statistics based on training data.
\cite{LID} extracted the local distance features and \cite{MD} imported Mahalanobis distance into the detection process.
Furthermore, \cite{DKNN,KNN,AbusnainaWAWWYM21} utilized the latent data neighbors to defend the attacks, while \cite{JTLA} incorporated feature statistics between layers to detection.
However, these statistics-based methods lack consideration of the dynamic features of inputs.
While \cite{ODD,WojcikMSKST21} utilized noise addition to accomplish the detection task.
\cite{Tian0LD21} applied wavelet transformation, converting features from the spatial domain into the frequency domain.
And recently, \cite{EPS} utilized a diffusion model to augment the features of AEs.
These variational analytical methods considered the dynamic features based on empirical experience, resulting in unreliable performance in unknown attacks.

Our proposed method(DSS) falls into the variational analytical category, as the constructed dynamically stable system involves disrupting and restoring inputs.
With the system theoretically showing the differences between AEs and NEs, our method exhibits good performance in AEs detection.
\section{Preliminary}
\noindent\textbf{AEs Generation Process}:
For a deeper comprehension of the AEs, we take the PGD \cite{PGD} attack as an example to show the AEs generation process, as follows:
\begin{gather}
    x_{t+1} = Clip_{x,\epsilon}(x_t + \alpha \cdot sign({{\nabla }_{x}}J(\theta, {{x}_{t}},y_0))) \label{eq:PGD_eq}
\end{gather}
In the equation \eqref{eq:PGD_eq}, $x_0=x$ is the initial condition while $x'=x_n$ is the ending result where $n$ is the total attacking times, $\theta$ is the parameter of the target model $f(\cdot)$, and $y_0$ is the true label of $x$.
As Eq.\eqref{eq:PGD_eq} shows, the input $x_t$ at $t$ loop is modified by the gradient descents of loss function $J(\cdot)$ with $\alpha$ step size.

\noindent\textbf{The Lyapunov Stable Theory}: For a system to be Lyapunov stable, it should be able to maintain stability at $x_0$ under various initial conditions.
The stability of the system is represented by a function $V(x)$, and if $V(x)>0,\dot{V}(x)\leq0$, the system is considered Lyapunov stable while $\dot{V}(x)$ represents the derivative of $V(x)$ with respect to perturbations.

\section{Proposed Method}\label{sec:4}
In Section \ref{sec:4}, we initially present our motivation. 
Then, the example stability mechanism is proposed and proven, and the dynamic stable system is proposed and conducted based on the proposed example stability mechanism. 
Finally, the running process of the proposed DSS to detect AEs is presented.

\subsection{Motivation}\label{sec:motivation}
Currently, detection methods can be broadly categorized into two types: 
those based on data statistics, which primarily analyze features from intermediate and logit layers of the model; 
and those based on variational analysis, which induce transformations to the inputs.
Statistics methods calculate the distance between inputs and prior data, easily leading to overfitting to the prior data and resulting in weak performance in generalization.
Additionally, existing variational analytical methods leverage the dynamic features of inputs, but these methods are conducted based on empirical experience and lack theoretical evidence, resulting in unreliable performance in unknown attacks. 

Therefore, leveraging the Lyapunov stability theory, we establish a \textbf{dynamically stable system} to generate the stability differences between NEs and AEs.
Furthermore, we monitor the dynamical stability features of inputs through our monitor module to accomplish adversarial detection, providing a complement to previous methods which are based on empirical experience.

\subsection{The  Example Stability Mechanism}\label{sec:proof}
In this subsection, the example stability mechanism is proposed and proven based on the Lyapunov stable theory.

Inspired by the adversarial examples (AEs) generation process shown in Eq.\eqref{eq:PGD_eq}, it can be viewed as a perturbation process of a dynamic system, described as follows:
\begin{equation}
  \left\{ \begin{aligned}
    & \dot{x}=\alpha{{\nabla }_{x}}J(\theta ,x,y_0) \\ 
   & {{y}_{p}}=f(x+\dot{x}\cdot\triangle t) \\ 
  \end{aligned} \right.  
  \label{eq:sys_aes}
\end{equation}

In Eq.\eqref{eq:sys_aes}, $\dot{x}$ represents the derivative of $x$ with respect to the iterations, and $\triangle t$ represents the iteration interval.
With iteratively adding malicious gradient information, the outputs lead to $y_p \neq y_0$, in which $y_0$ and $y_p$ represent the true label and the predicted class.

Based on the dynamic system of adversarial examples generation,  the example stability mechanism can be proposed, in which the malicious gradient information  ${{\nabla }_{x}}J(\theta ,x,y_p)$ is used to perturb the inputs and a novel control term $u(t)$ is proposed to maintain the stability of normal original examples,  which can be represented as:
\begin{equation}
  \left\{ \begin{aligned}
    & \dot{x}=\alpha{{\nabla }_{x}}J(\theta ,x,y_p)+u(t) \\ 
   & {{y}_{p}}=f(x+\dot{x}\cdot\triangle t) \\ 
  \end{aligned} \right.
  \label{eq:sys_state}
\end{equation}

Additionally, in our example stability mechanism, the quadratic Lyapunov function (i.e., $V(x)=(x-x_0)^2$) is introduced to represent the stability status of input examples. 
Obviously, when $x$ is equal to $x_0$ (i.e., $x=x_0$), the stability status of example $x$ will be 0 (i.e., $V(x)=0$) , and when $x$ is not equal to $x_0$ (i.e., $x\ne x_0$), the stability status of example $x$ will be larger than 0 (i.e., $V(x)>0$) satisfying the initial Lyapunov stable condition $V(x)>0$. 
What's more, to ensure the stability of $x$ at $x_0$, there should also be $\dot{V}(x)\le 0$, i.e.,
\begin{equation}
  \begin{aligned}
    & \dot{V}(x)\le 0 \\ 
    & \Rightarrow 2(x-{{x}_{0}})\cdot \dot{x} \le 0 \\ 
   & \Rightarrow 2(x-{{x}_{0}})\cdot [\alpha {{\nabla }_{x}}J(\theta ,x,y_p)+u(t)]\le 0 \\ 
  \end{aligned}
  \label{eq:h}
\end{equation}
Due to the uncertainty of $2(x-x_0)$ as it represents the image disturbance, it is necessary to satisfy $\alpha {{\nabla }_{x}}J(\theta ,x,y)+u(t)=0$ to ensure $\dot{V}(x)\le 0$. 
Hence, we set $u(t)=-\alpha {{\nabla }_{x}}J(\theta ,x,y_p)$ to satisify $\dot{V}(x)=0$.

Therefore, in our example stability mechanism, the proposed control term $u(t)$ is set as $-\alpha {{\nabla}_{x}}J(\theta ,x,y_p)$, i.e., $u(t)=-\alpha {{\nabla}_{x}}J(\theta ,x,y_p)$. 
By doing this, the control term $u(t)$ can make a normal example $x$ stable at the original example $x_0$,  i.e., $\dot{x}=\alpha{{\nabla }_{x}}J(\theta ,x,y_p)-\alpha{{\nabla }_{x}}J(\theta ,x,y_p)=0$. 
However, due to the adversarial noises are analogous to Gaussian noises for $u(t)$, the $u(t)$ will see the label of $x'$ as the original label $y_0$ while the perturbation is based on the malicious label $y_p\neq y_0$, resulting in $x'$ not stable at the initial status, i.e., $\dot{x}'=\alpha \cdot {{\nabla }_{x'}}J(\theta ,x',y_p)-\alpha{{\nabla }_{x'}}J(\theta,x',y_0)\ne 0$. Hence, adversarial examples will diverge, thereby deviating from the initial point.

\subsection{The Dynamically Stable System (DSS)}\label{sec:method}
\begin{figure}[htbp]
  \centerline{\includegraphics[width=0.5\textwidth]{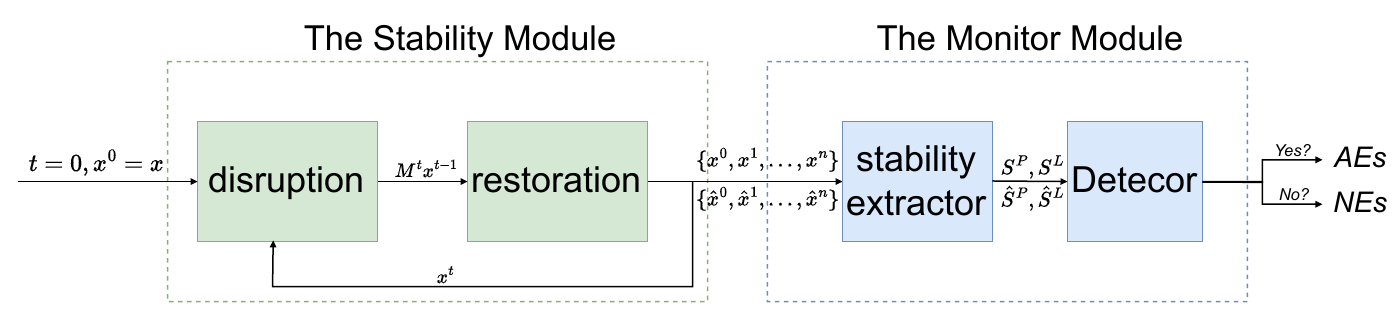}}
  \caption{The flowcharts of the dynamically stable system}
  \label{fig8}
  \vspace{-5pt}
\end{figure}

In this section, we introduce our Dynamically Stable System(DSS), which comprises the stability and monitor modules.
In the stability module, the repeated disruption and restoration actions are performed to present the stability of the inputs, 
while the monitor module distinguishes between AEs and NEs based on the different stability, as shown in Fig.\ref{fig8}.

\subsubsection{The Stability Module}
The stability module comprises two actions: disruption and restoration.
In the disruption action, we disrupt critical information of inputs based on the gradient descents.
In the restoration action, we repair the disrupted parts using the remaining information from the inputs.

In the disruption action, considering that the success of AEs is attributed to the addition of harmful noises on NEs, we contemplate disrupting these types of noises.
Previous works \cite{WangGZLQ021,ZhangWHHW0L22} have provided that the saliency map of inputs could be calculated by the gradient descents of the loss.
Moreover, considering that AEs alter the predicted class and the loss is calculated based on the predicted class, we focus on disrupting the elements that contribute to the minimum loss.
Assuming the system is at the $t^{th}$ loop, we calculate the gradient $G_t$ of the loss as follows:
\begin{equation}
  G_t ={{\nabla }_{x}}L_{CE}(x_{t-1},f(x_{t-1}))
  \label{eq:M}
\end{equation}
In Eq.\eqref{eq:M}, $f(x_{t-1})$ represents logits of inputs by the classifier $f(\cdot)$, and $L_{CE}$ represents the cross entrophy loss function.
With the gradient $G_t$, we sort the values of $G_t$ and select the value located at the disrupting ratio $r=3\%$ position as the disrupting threshold $\tau$.
Furthermore, values in $G_t$ greater than $\tau$ are retained, while values less than $\tau$ are disrupted.
Then the disrupting matrix $M_t$ can be calculated as follows:
\begin{align}
  M_{t,i,j} = \begin{cases}
      1,\quad if \ {G_{t,i,j}}>{{\tau }} \\ 
     0,\quad if \ {G_{t,i,j}}\leq{{\tau }} \\ 
  \end{cases}
  \label{eq:matrix}
\end{align}
In Eq.\eqref{eq:matrix}, the notations $i$ and $j$ denote the row and column coordinates of the disrupted pixel, respectively. 
With the disrupting matrix $M_t$, we get the disrupted data $M_{t}x_{t-1}$.

\begin{algorithm}[tbhp]
  \caption{Dynamically stable system}
  \label{alg:algorithm}
  \textbf{Input}: Clean set $X_{clean}$, noisy set $X_{noisy}$, adversarial set $X_{adv}$\\
  \textbf{Parameter}: Classifier $f(\cdot)$, inpainting model $g(\cdot)$, length of set $N$, loop times $n$\\
  \textbf{Output}: Dynamically stable feature matrices $S^{P}$, $\hat{S}^{P}$ in pixel-wise, $S^{L}$, $\hat{S}^{L}$ in logit-wise
  \begin{algorithmic}[1]
    \STATE $i=1$
      \FOR{ $x$ in [$X_{clean},X_{noisy},X_{adv}$]}
        \STATE Let $t=1$, $x_{0} = x$.
        \WHILE{$t<n+1$}
        \STATE Calculate masks $M_t$ by backpropagating the outputs of $f(x_{t-1})$ via Eqs.\eqref{eq:M} and \eqref{eq:matrix}
        \STATE Get the generated examples $\hat{x}_t$ via Eq.\eqref{eq:zt}
        \STATE Get the composed examples $x_t$ via Eq.\eqref{eq:xt_generate}  
        \STATE Extract stability features $l_{t,1,p}$, $\hat{l}_{t,1,p}$ and $l_{t,2,p}$, $\hat{l}_{t,2,p}$ via Eqs.\eqref{eq:l_1} and \eqref{eq:l_2}
        \STATE Record the features: $S^{P}_{i,t}=l_{t,1,p}$, $\hat{S}^{P}_{i,t}=\hat{l}_{t,1,p}$, $S^{L}_{i,t}=l_{t,2,p}$, $\hat{S}^{L}_{i,t}=\hat{l}_{t,2,p}$
        \STATE $t = t + 1$  
        \ENDWHILE
        \STATE $i = i + 1$
      \ENDFOR
    \STATE \textbf{return} Dynamically stable features $S^P$, $\hat{S}^P$, $S^L$, $\hat{S}^L$
  \end{algorithmic}
\end{algorithm}

In the restoration action, we employ an inpainting model to repair the disrupted data $M_{t}x_{t-1}$.
The inpainting model $g(\cdot)$ takes two inputs: the disrupted data $M_{t}x_{t-1}$ and the disrupting matrix $M_t$.
The disrupted data  $M_{t}x_{t-1}$ provides initial information for the model, while the disrupting matrix $M_t$ identifies information positions to take.
The outputs of the model are generated examples $\hat{x}_t$ with the same shape as inputs, and the process is as follows:
\begin{equation}
  \hat{x}_t = g(M_{t}x_{t-1}, M_t)
  \label{eq:zt}
\end{equation}
Using the generated examples $\hat{x}_t$, we can calculate the outputs at the $t^{th}$ loop as follows:
\begin{equation}
  {{x}_{t}}=(1-M_t)\cdot \hat{x}_t+M_tx_{t-1}
  \label{eq:xt_generate}
\end{equation}
With the composed examples $x_t$, we can proceed to the $(t+1)^{th}$ loop to calculate $x_{t+1}$.

Our stability module is built on the disruption and restoration actions described above.
The state-space equation of our module is as follows:
\begin{equation}
  \label{eq:Lyapunov_function}
  \left\{ \begin{aligned}
   & \dot{x_t }=x_t-x_{t-1}=(1-M_t)[g(M_{t}x_{t-1},M_t)-x_{t-1}] \\ 
   & L_t={{L}_{CE}}({x_t} ,f(x_t))-{{L}_{CE}}(x_{t-1},f(x_{t-1})) \\
  \end{aligned} \right.
\end{equation}
In Eq.\eqref{eq:Lyapunov_function}, the variables at the $t^{th}$ loop of the dynamically stable system are presented, in which $L_t$ represents the state variable of the system.
Taking the initial inputs $x_0=x$, we iterate the disrupting and restoring actions total of $n$ times, with the outputs of each iteration serving as the inputs for the next one, getting the module outputs $\{x_1,x_2,...,x_n\}$ and $\{\hat{x}_1,\hat{x}_2,...,\hat{x}_n\}$.

\subsubsection{The Monitor Module}

Through the stability module, we obtain the states $\{x_1,x_2,...,x_n\}$ and $\{\hat{x}_1,\hat{x}_2,...,\hat{x}_n\}$ of the input after repeated disruption and restoration. 
In the monitor module, the system consists of two parts: the stability extractor and the detector.
In the stability extractor, it extracts the stability as follows:
\begin{equation}
  \label{eq:l_1}
  \left\{ \begin{aligned}
    & l_{t,1,p}={{\left\| {{x}_{t}}-{{x}_{0}} \right\|}_{p}} \\
    & \hat{l}_{t,1,p}={{\left\| {{\hat{x}}_{t}}-{{x}_{0}} \right\|}_{p}}
  \end{aligned} \right.
\end{equation}
In the formula above, $p$ denotes the chosen norm type. 
$l_{t,1,p}$ and $\hat{l}_{t,1,p}$ denote detecting features at $t_{th}$ loop in pixel-wise, while $l_{t,2,p}$ and $\hat{l}_{t,2,p}$ denote detecting features in logit-wise which could be seen as below:
\begin{equation}
  \label{eq:l_2}
  \left\{ \begin{aligned}
    & l_{t,2,p}={{\left\| f({{x}_{t}})-f({{x}_{0}}) \right\|}_{p}} \\
    & \hat{l}_{t,2,p}={{\left\| f({{\hat{x}}_{t}})-f({{x}_{0}}) \right\|}_{p}} \\
  \end{aligned} \right.
\end{equation}
What's more, with stability features $l_{t,1,p}, \hat{l}_{t,1,p}, l_{t,2,p}, \hat{l}_{t,2,p}$ extracted, we utilize a detector based on logistic regression to accomplish the detection.
The detector will output a score, where the score closer to 1 indicates a higher confidence that the input is an adversarial example, while the score closer to 0 suggests a higher likelihood that the input is a normal example.

\subsection{The Detection Process of our DSS}\label{sec:process}

With our DSS introduced, we can get the whole detection process as Algorithm \ref{alg:algorithm} shows.
Through the system, we obtain the stability features $l_t$ at different iterations and then accomplish detection tasks based on logistic regression.

In the algorithm \ref{alg:algorithm}, 
line 1 sets $i=1$ to record the input number. 
Line 2 selects one example $x$ as the input for the system from the sets of clean, noisy and adversarial examples.
Line 3 sets $t=1$ as the initial iteration number, with $x_0=x$ as the input of the first iteration.
Line 4 specifies the total iterations of the system as $n$.
Line 5 inputs $x_{t-1}$ to the classifier $f(\cdot)$ and backpropagates the classifier logits to calculate the disrupting matrix $M_t$ by equation \eqref{eq:matrix}.
Line 6 inputs the disrupted data $M_tx_{t-1}$ and mask $M_t$ into the inpainting model $g(\cdot)$ to generate the generated example $\hat{x}_t$ by equation \eqref{eq:zt}.
Line 7 calculates the composed example $x_{t}$ with the equation\eqref{eq:xt_generate}.
Line 8 extracts the stability features $l_{t,1,p}$, $\hat{l}_{t,1,p}$, $l_{t,2,p}$, and $\hat{l}_{t,2,p}$ by the equations \eqref{eq:l_1} and \eqref{eq:l_2}.
Line 9 records the stability features into the matrices $S^P$, $\hat{S}^P$, $S^L$, and $\hat{S}^L$.
Lines 10-13 end the loop of the system and the inputs traversal, respectively.
Using these features, we train a logistic regressor to perform detection on the inputs.

\begin{figure}[htbp]
  \centerline{\includegraphics[width=0.5\textwidth]{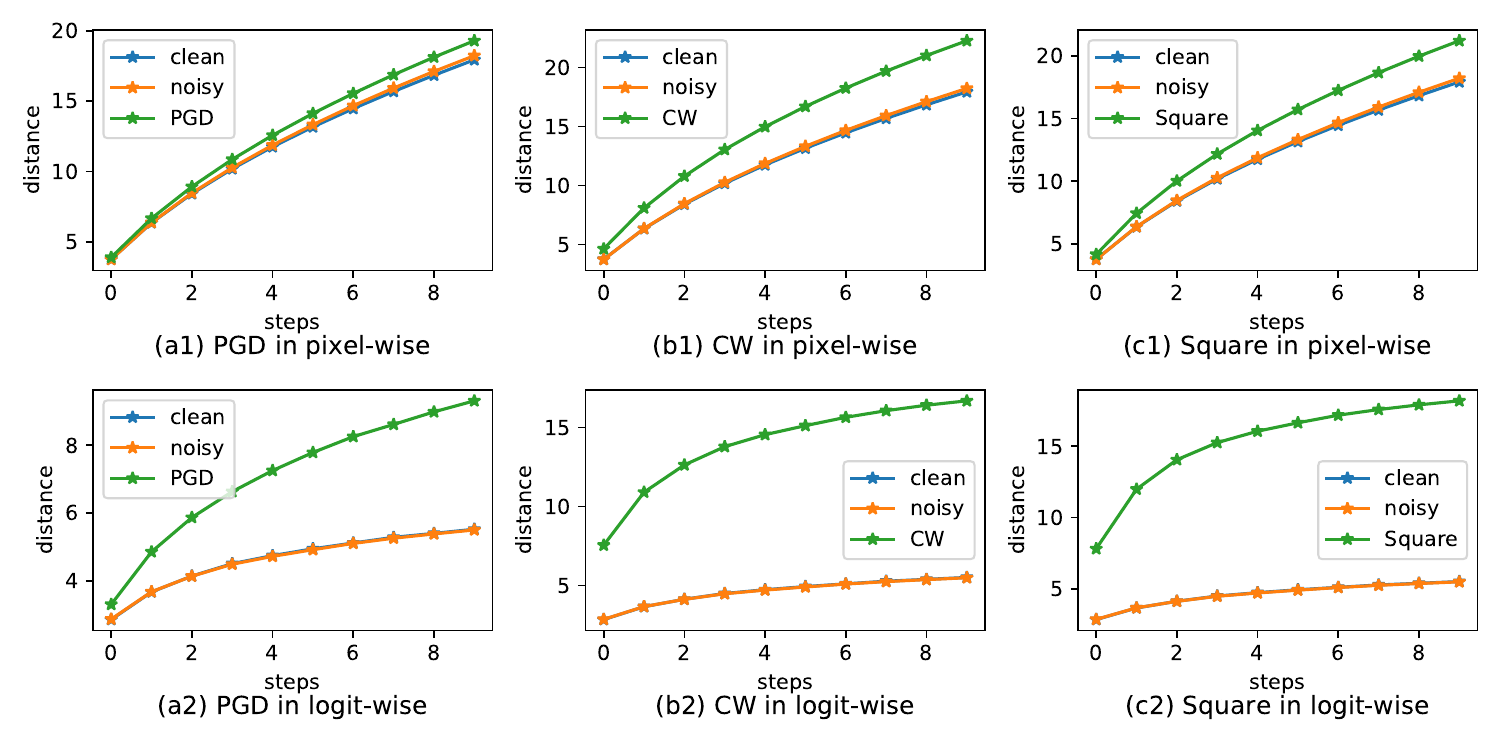}}
  \vspace{-10pt}
  \caption{Distance of composed examples and original examples at each loop.}
  \vspace{-10pt}
  \label{fig6}
\end{figure}

Furthermore, we validate the rationality of stability in our method as shown in Fig.\ref{fig6}.
In Fig.\ref{fig6}, we calculate the mean distance between inputs $x$ and composed data $x_t$ under CIFAR10 \& VGG19 with PGD, CW and Square attacks, and it is evident that AEs exhibit higher stability changes compared to clean and noisy examples both at the pixel and logit levels.
\section{Experiments}
\subsection{Experiment Setup}
\subsubsection{Datasets}
In this paper, three benchmarks are utilized to verify the effectiveness of our proposed method, namely MNIST \cite{lecun1998gradient}, CIFAR10 \cite{cifar10} and CIFAR100. 
\subsubsection*{Classifier Models}
For CIFAR10 and CIFAR100, we select VGG19 and ResNet50 networks as validation models. 
For MNIST, the LeNet network is chosen.

\subsubsection*{Inpainting Models}
Due to the irregular positions that need restoring, existing inpainting models primarily restore information at fixed locations.
Therefore, we use the partial convolutional layers \cite{pcov,LiuRSWTC18} due to their capability to restore irregular positions. 

\subsubsection*{Attacks Setup}
We employ six types of attacks for evaluation, namely FGSM, PGD, CW, DeepFool, Square, and AutoAttack.
The attack Square is a black-box attack while the others are gray-box attacks.
Additionally, they can be divided into gradient-based(FGSM, PGD, AutoAttack) and optimization-based(the others).
In terms of norm type, we categorize them as $l_2$(CW and DeepFool) and $l_\infty$(other four attacks).
For parameter settings, we choose attack intensity $\epsilon=8/255$ for CIFAR10 \& CIFAR100, $\epsilon=0.3$ for MNIST, step size $\alpha=\frac{\epsilon}{4}$, CW optimization $steps=500$, with the remaining parameters set to the default by the torchattacks library.

\subsubsection*{Metric}
Like the majority of existing works, we utilize the area under roc curve(AUC) \cite{prandroc} as the performance metric to compare these methods.

\subsubsection*{Baseline}
In our experiment, the examples conclude three parts: clean examples, noisy examples, and adversarial examples. 
The inclusion of noisy examples is due to the consideration that noisy data is commonly present in the natural environment and typically does not pose misclassification to classifier models. 
In the experiment, we select examples that the classifier successfully classify but fail in corresponding AEs.
For noisy examples, random noises are added to clean examples, with the same intensity of AEs.
Then, the detection features extracted by existing detection methods are split into an 80-20 ratio, with 80$\%$ of the features used to train a logistic regressor and 20$\%$ used for testing the detection performance. 

\begin{table*}[htbp]
  \caption{Detection performance(AUC score(\%)) comparison on CIFAR10 and CIFAR100.The bold values represent the best detection performance, the underlined ones represent the second-best detection performance, and subscript values denote the detection performance combined with our method.}
  \centering
  \resizebox{1.0\textwidth}{!}{
    \begin{tabular}{cccccccccccccc}
      \hline
      \scalebox{1.7}{\phantom{H} }    & \multicolumn{6}{c}{CIFAR10\&VGG19}                         &  & \multicolumn{6}{c}{CIFAR10\&ResNet50}            \\ 
      \hline
            & FGSM   & PGD    & CW     & DeepFool & Square & AutoAttack &  & FGSM & PGD & CW & DeepFool & Square & AutoAttack \\
      FS    & 0.7957\textsubscript{0.9675}  & 0.9769\textsubscript{0.9978}& 0.9109\textsubscript{0.9755} & \underline{0.9133}\textsubscript{0.9612}  & 0.8847\textsubscript{0.9726} & \underline{0.9768}\textsubscript{0.9939}           & &0.8936\textsubscript{0.9938}&\underline{0.9953}\textsubscript{1.0000}&0.8882\textsubscript{0.9841}&0.8439\textsubscript{0.9636}&0.7996\textsubscript{0.9617}&\underline{0.9957}\textsubscript{0.9999}\\
      KDBU  & 0.7570\textsubscript{0.9663}  & 0.9789\textsubscript{0.9985}& \underline{0.9235}\textsubscript{0.9793} & 0.8514\textsubscript{0.9526}  & 0.8724\textsubscript{0.9701} & 0.9433\textsubscript{0.9949}                       & &0.8719\textsubscript{0.9920}&0.9946\textsubscript{0.9999}&\underline{0.9480}\textsubscript{0.9821}&\underline{0.9663}\textsubscript{0.9833}&\underline{0.9683}\textsubscript{0.9822}&0.9918\textsubscript{0.9998}\\
      LID   & \underline{0.9612}\textsubscript{0.9820}  & 0.9478\textsubscript{0.9980}& 0.9218\textsubscript{0.9764} & 0.8452\textsubscript{0.9577}  & 0.9108\textsubscript{0.9770} & 0.9072\textsubscript{0.9973}                       & &0.9398\textsubscript{0.9943}&0.9577\textsubscript{0.9999}&0.8123\textsubscript{0.9828}&0.7262\textsubscript{0.9624}&0.7414\textsubscript{0.9642}&0.9625\textsubscript{0.9998}\\
      MD    & 0.9458\textsubscript{0.9771}  & \textbf{0.9954}\textsubscript{0.9995}& 0.8700\textsubscript{0.9720} & 0.8051\textsubscript{0.9519}  & 0.8555\textsubscript{0.9718} & 0.9504\textsubscript{0.9980}                          & &\textbf{0.9936}\textsubscript{0.9987}&0.9509\textsubscript{0.9999}&0.9134\textsubscript{0.9848}&0.7970\textsubscript{0.9637}&0.8763\textsubscript{0.9674}&0.9790\textsubscript{0.9999}\\
      JTLA  & 0.7206\textsubscript{0.9671}  & 0.5732\textsubscript{0.9944}& 0.8908\textsubscript{0.9802} & 0.8362\textsubscript{0.9605}  & \underline{0.9666}\textsubscript{0.9866} & 0.5852\textsubscript{0.9908}                       & &0.8313\textsubscript{0.9931}&0.6834\textsubscript{0.9999}&0.9382\textsubscript{0.9912}&\textbf{0.9814}\textsubscript{0.9897}&\textbf{0.9775}\textsubscript{0.9868}&0.7050\textsubscript{0.9998}\\
      LiBRe & 0.7481\textsubscript{0.9670}  & 0.6316\textsubscript{0.9939}& 0.8315\textsubscript{0.9728} & 0.8583\textsubscript{0.9570}  & 0.8365\textsubscript{0.9721} & 0.6514\textsubscript{0.9910}                                   & &0.7831\textsubscript{0.9919}&0.5814\textsubscript{0.9999}&0.8833\textsubscript{0.9844}&0.9018\textsubscript{0.9668}&0.9224\textsubscript{0.9660}&0.5583\textsubscript{0.9998}\\
      EPS   & 0.7682\textsubscript{0.9662}  & 0.7624\textsubscript{0.9939}& 0.9054\textsubscript{0.9886} & 0.5836\textsubscript{0.9518}  & 0.5116\textsubscript{0.9711} & 0.8387\textsubscript{0.9955}                                   & &0.6640\textsubscript{0.9919}&0.8186\textsubscript{1.0000}&0.7967\textsubscript{0.9846}&0.5268\textsubscript{0.9599}&0.5033\textsubscript{0.9592}&0.9598\textsubscript{0.9999}            \\
      DSS(ours)  &\textbf{0.9661}  & \underline{0.9939}& \textbf{0.9716} & \textbf{0.9517}  & \textbf{0.9705} & \textbf{0.9910}                                                                                                               & &  \underline{0.9919}&\textbf{0.9999}&\textbf{0.9822}&0.9597&0.9592&\textbf{0.9998}            \\ \hline
      \scalebox{1.7}{\phantom{H} }     & \multicolumn{6}{c}{CIFAR100\&VGG19}                       &  & \multicolumn{6}{c}{CIFAR100\&ResNet50}           \\ \hline
            & FGSM   & PGD    & CW     & DeepFool & Square & AutoAttack &  & FGSM & PGD & CW & DeepFool & Square & AutoAttack \\
      FS    & 0.7559\textsubscript{0.9477}  & 0.9669\textsubscript{0.9981}&  0.8091\textsubscript{0.9424}& 0.7961\textsubscript{0.8873}  & 0.7744\textsubscript{0.9183} & 0.9703\textsubscript{0.9974}                                                     & & 0.9521\textsubscript{0.9910}&0.9902\textsubscript{0.9997}&0.8222\textsubscript{0.9262}&0.7102\textsubscript{0.8687}&0.7443\textsubscript{0.9040}&0.9919\textsubscript{0.9998} \\
      KDBU  & 0.8037\textsubscript{0.9479}  & 0.9509\textsubscript{0.9977}&  0.8751\textsubscript{0.9425}& 0.8703\textsubscript{0.8822}  & 0.7678\textsubscript{0.9166} & 0.9393\textsubscript{0.9961}                                                     & & 0.9135\textsubscript{0.9846}&\underline{0.9910}\textsubscript{0.9995}&0.8891\textsubscript{0.9434}&0.8593\textsubscript{0.9031}&0.8631\textsubscript{0.9158}&0.9881\textsubscript{0.9995}\\
      LID   & 0.9419\textsubscript{0.9739}  & 0.9338\textsubscript{0.9984}&  0.8403\textsubscript{0.9453}& 0.6989\textsubscript{0.8846}  & 0.7581\textsubscript{0.9165} & 0.9395\textsubscript{0.9982}                                                     & & 0.9151\textsubscript{0.9899}&0.9369\textsubscript{0.9996}&0.7042\textsubscript{0.9221}&0.5896\textsubscript{0.8627}&0.6974\textsubscript{0.9022}&0.9377\textsubscript{0.9995}\\
      MD    & \textbf{0.9913}\textsubscript{0.9924}  & \underline{0.9862}\textsubscript{0.9987}&  \underline{0.9288}\textsubscript{0.9608}& 0.8703\textsubscript{0.9148}  & \underline{0.9154}\textsubscript{0.9464} & \underline{0.9880}\textsubscript{0.9985}           & & \textbf{0.9974}\textsubscript{0.9993}&0.9881\textsubscript{0.9998}&\underline{0.8963}\textsubscript{0.9464}&0.7858\textsubscript{0.8914}&0.8777\textsubscript{0.9301}&\underline{0.9940}\textsubscript{0.9993}\\
      JTLA  & 0.7904\textsubscript{0.9486}  & 0.6545\textsubscript{0.9973}&  0.8889\textsubscript{0.9538}& \underline{0.8794}\textsubscript{0.9175}  & 0.9149\textsubscript{0.9416} & 0.6538\textsubscript{0.9960}                             & & 0.8307\textsubscript{0.9876}&0.7066\textsubscript{0.9995}&0.9029\textsubscript{0.9603}&\textbf{0.9344}\textsubscript{0.9517}&\textbf{0.9052}\textsubscript{0.9418}&0.7446\textsubscript{0.9994}\\
      LiBRe & 0.7976\textsubscript{0.9483} & 0.5374\textsubscript{0.9971}&  0.8022\textsubscript{0.9410}& 0.8449\textsubscript{0.8816}  & 0.8028\textsubscript{0.9167} & 0.5338\textsubscript{0.9960}                                                      & & 0.8058\textsubscript{0.9848}&0.7385\textsubscript{0.9994}&0.8265\textsubscript{0.9324}&0.8190\textsubscript{0.8653}&0.8135\textsubscript{0.8938}&0.6657\textsubscript{0.9994}\\
      DSS(ours)  & \underline{0.9469} &\textbf{0.9970} &\textbf{0.9408}  &\textbf{0.8815}&\textbf{0.9159}  &\textbf{0.9960}                                                                                                                                   & &\underline{0.9846} &\textbf{0.9994}&\textbf{0.9202}&\underline{0.8626}&\underline{0.8921}&\textbf{0.9994}\\ \hline
      \end{tabular}
  }   
  \label{tab:grey_cifar}%
  \end{table*}

We compare our proposed method with seven other detection methods, namely KDBU \cite{KDBU}, FS \cite{FS}, LID \cite{LID}, MD \cite{MD}, JTLA \cite{JTLA}, LIBRE \cite{LIBRE}, and EPS \cite{EPS}. 
The parameters of these methods are either taken from the respective papers or source codes. 
And for the EPS detection method, it is only compared on CIFAR10 since the method requires a diffusion model, but the authors only provided the diffusion model for CIFAR10.

\subsection{Detection Performance Results}

In this subsection, we compare our method with seven other detection approaches, quantifying the detection performance by the AUC metric. 
As shown in Table.\ref{tab:grey_cifar}, we present the detection results of the eight methods on CIFAR10 and CIFAR100, with both VGG19 and ResNet50 as target models, respectively. 
In Table.\ref{tab:grey_cifar}, the bold values represent the best detection performance; the underlined ones represent the second-best detection performance; and subscript values denote the detection performance combined with our method.

From Table.\ref{tab:grey_cifar}, it is evident that our DSS method outperforms other seven detection approaches. 
Our approach achieves AUC values of 97.81\% and 94.47\% on CIFAR10 and CIFAR100, respectively, surpassing the SOTA of other methods with AUC values of 91.10\%(MD) and 93.49\%(MD). 
Additionally, to demonstrate the compatibility of the features extracted by our method, we combine our features with those of others.
The results of this combined manner are presented as the subscript values in Table.\ref{tab:grey_cifar}.
It shows that our method exhibits good compatibility with other approaches after feature combination, demonstrating a general upward trend in AUC values.

\begin{table}[htbp]
  \vspace{-5pt}
  \caption{Detection ability (AUC score(\%)) comparison on MNIST}
  \centering
  \resizebox{1.0\columnwidth}{!}{
        \begin{tabular}{ c c c c c c c}
        \hline
        \multicolumn{7}{c}{MNIST \& LeNet}  \\
        \hline
        & FGSM    & PGD   & CW    & DeepFool & Square & AutoAttack \\
  
        FS  & 0.7507  & 0.8096& 0.8833 & 0.7484  & 0.7418 & \underline{0.9720}                                       \\
        
        KDBU& 0.9899  & \underline{0.9796}& 0.9748 & 0.9838  & 0.9934 & 0.7701                                            \\
        
        LID & 0.9260  & 0.8912& 0.8321 & 0.8513  & 0.9378 & 0.9285                                            \\
        
        MD  & \underline{0.9945}  & 0.9684& 0.8923 & 0.8785  & 0.9884 & 0.9688                                          \\
        
        JTLA& 0.9687  & 0.9744& \textbf{0.9981} & \textbf{0.9987}  & 0.9874 & 0.8217                                                \\
  
  
        LiBRe& 0.9861 & 0.9654& 0.9749 & 0.9762  & \underline{0.9969} & 0.9412                                                 \\
  
  
        DSS(ours) & \textbf{1.0000}  &\textbf{1.0000} &\underline{0.9911}  &\underline{0.9985}   &\textbf{1.0000}  &\textbf{1.0000}                                    \\
        \hline
        \end{tabular}%
  }  
  \label{tab:grey_mnist}%
  \end{table}

In Table.\ref{tab:grey_mnist}, we present the detection results of our method and other approaches under the MNIST \& LeNet condition. 
The table shows that our method is the second-best only under optimization-based attacks (CW and DeepFool) and is the best under the remaining attack strategies. 
Our method achieves an average AUC of 99.83\%, surpassing other methods' SOTA with an average AUC of 97.35\% (LIBRE).

\subsection{Detection Performance Comparison Under Different Attack intensities}
In this subsection, we demonstrate the detection capabilities of our method and other approaches under different attack intensities. 
We take PGD(gradient-based) and CW(optimization-based) attacks as examples, conducted under the CIFAR10 \& VGG19 condition. 
For the PGD($l_\infty$ norm) attack, the attack strength $\epsilon$ ranges from [1/255, 2/255, 4/255, 6/255, /8/255, 12/255, 16/255], and the step size $\alpha=\frac{\epsilon}{4}$; 
for the CW($l_2$ norm) attack, the optimization $steps$ ranges from [10, 20, 50, 100, 500, 3000, 10000]. 

\begin{figure}[htbp]
  \centerline{\includegraphics[width=0.5\textwidth]{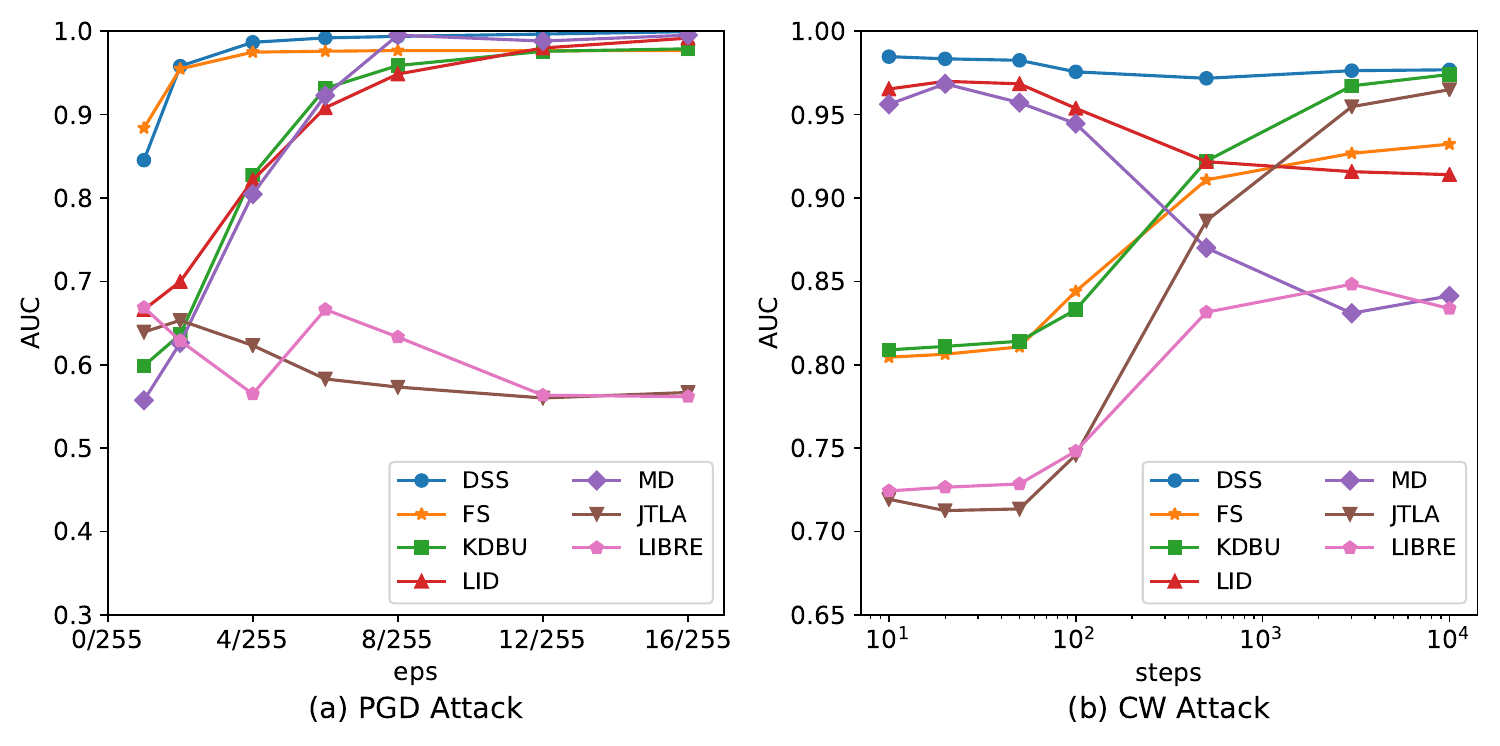}}
  \vspace{-10pt}
  \caption{Detection performance comparison (AUC score(\%)) under different attack intensities.}
  \vspace{-10pt}
  \label{fig4}
\end{figure}

\begin{figure*}[htbp]
  \centerline{\includegraphics[width=1.0\textwidth]{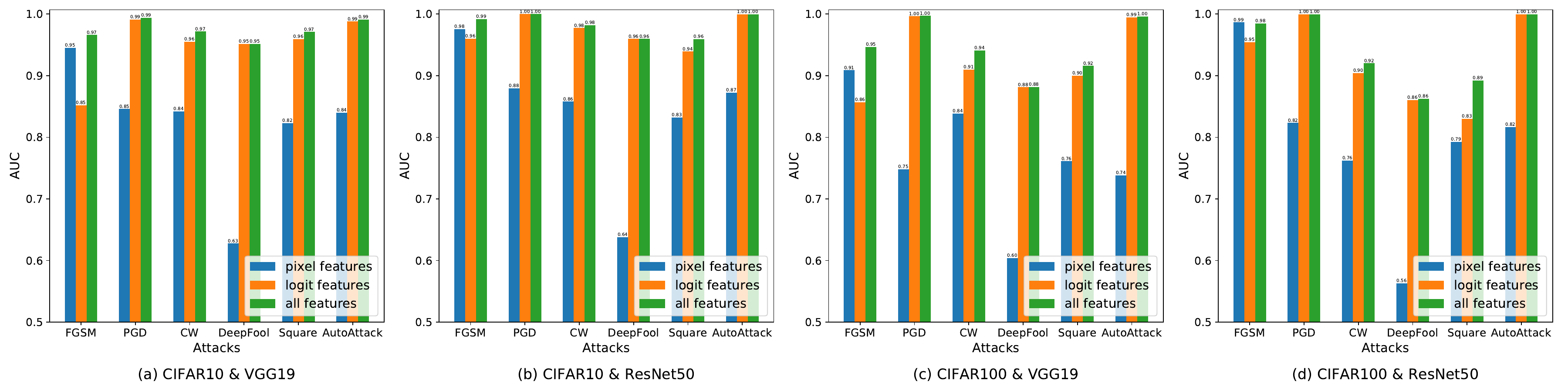}}
  \vspace{-10pt}
  \caption{Ablation study. The blue bars illustrate the detection performance of pixel-wise DSS features, the orange bars represent the detection performance of logit-wise DSS features, and the green bars showcase the overall detection performance of our DSS method.}
  \label{fig5}
  \vspace{-5pt}
\end{figure*}

From Fig.\ref{fig4}, we observe that under PGD($l_\infty$ norm) attack, our method performs slightly worse than FS at $\epsilon=1/255$, but it outperforms the other methods;
when $\epsilon\in[2/255, 4/255, 6/255]$, our method exhibits the best detection performance;
when $\epsilon\in[8/255, 12/255, 16/255]$, our method, along with FS, KDBU, LID, and MD, demonstrate outstanding detection performance. 
Under CW($l_2$ norm) attack, due to the large span of our x-axis, we use a logarithmic scale.
LID and MD methods decrease notably with increasing optimization $steps$, while FS, KDBU, JTLA, and LIBRE exhibit weak detection performance at lower optimization $steps$. 
In contrast, our method's detection performance remains relatively stable and consistently high as the optimization $steps$ change, indicating that the dynamic features introduced in our method exhibit more stable detection performance.

\subsection{Generalization Study}
Considering that we often face the challenges of unknown attacks in the real world, therefore, 
we assess the generalization performance of our method by using a detector trained on FGSM attack to detect other attacks.
Besides, to validate the effectiveness of our method features in pixel-wise and logit-wise, we conduct the generalization study on these two types of features labeled as $P$(pixel-wise) and $L$(logit-wise), respectively.

\begin{table}[htbp]
  \caption{Generalization comparison (AUC score(\%)) on CIFAR10}
  \centering
  \resizebox{1.0\columnwidth}{!}{
        \begin{tabular}{ c c c c c c c c}
        \hline
                &&PGD   & CW    & DeepFool & Square & AutoAttack & Avg \\ 
      \hline
        FS      &&0.9426&0.9050&\underline{0.9055}&0.8464&\textbf{0.9573} &   \underline{0.9114}                       \\
        
        KDBU    &&0.0273&0.8560&0.8514&0.8723&0.0699    &   0.5354                          \\
        
        LID     &&0.8265&0.8758&0.6933&0.6843&0.8659   &  0.7892                               \\
        
        MD      &&0.2983&0.8934&0.7605&0.7151&0.2950      &   0.5925                   \\
        
        JTLA    &&0.4779&0.8735&0.8176&\underline{0.9478}&0.5216    &  0.7277                                    \\
  
  
        LiBRe   && 0.4800 & 0.8151 & 0.8457 & 0.8209 & 0.4757 & 0.6875                                   \\
 
        EPS     && 0.7634 & 0.9042 & 0.4164 & 0.5116 & 0.8387 & 0.6868                                     \\
  
        DSS(P)(ours)   && 0.7955 & 0.8016 & 0.5460 & 0.7471 & 0.7965 & 0.7373 \\

        DSS(L)(ours)   && \underline{0.9558} & 0.9020 & 0.8851 & 0.8558 & 0.9546 & 0.9107                         \\

        DSS(L+P)(ours) && 0.9435 & \underline{0.9116} & 0.7529 & 0.8778 & 0.9417 & 0.8855 \\


        DSS(L)+FS+JTLA(ours) && \textbf{0.9565} & \textbf{0.9599} & \textbf{0.9462} & \textbf{0.9538} & \underline{0.9561}  & \textbf{0.9545} \\
        \hline
        \end{tabular} 
  }
  \label{tab:generalization}
  \end{table}

As shown in Table.\ref{tab:generalization}, under the CIFAR10\&VGG19 condition, our proposed method achieves a generalization accuracy of 91.07\% at the logit level, slightly lower than the generalization performance of FS 91.14\%. 
In contrast, the remaining methods, KDBU, MD, JTLA, LIBRE, and EPS, exhibit a generalization AUC of less than 0.5. 
Additionally, combining our method with FS and JTLA, which involves the fusion of static and dynamic features, results in an average generalization AUC of 95.45\%.

\subsection{Ablation Study}
We utilized logit-wise and pixel-wise features in the previous subsection to accomplish adversarial detection tasks. 
To validate their effectiveness, in this subsection, we conduct adversarial detection at pixel-wise and logit-wise levels, separately. 
Similar to the experiments in Table.\ref{tab:grey_cifar}, we perform detection on VGG19 and ResNet50 architectures on CIFAR10 and CIFAR100.
As shown in Fig.\ref{fig5}, features at the pixel-wise level demonstrate better detection performance against FGSM attack, while features at the logit-wise level exhibit superior detection performance against PGD, CW, DeepFool, Square, and AutoAttack attacks.

\subsection{Sensitivity Study}
For the reason that  our DSS method involves a disruption phase, the disrupting ratio $r$ is crucial in our method. 
In this subsection, under the CIFAR10 \& CIFAR100 of ResNet50 condition, we validate the detection performance of our experiments under different disrupting ratios to further affirm the effectiveness of our approach. 
\begin{figure}[htbp]
  \centerline{\includegraphics[width=1.05\columnwidth]{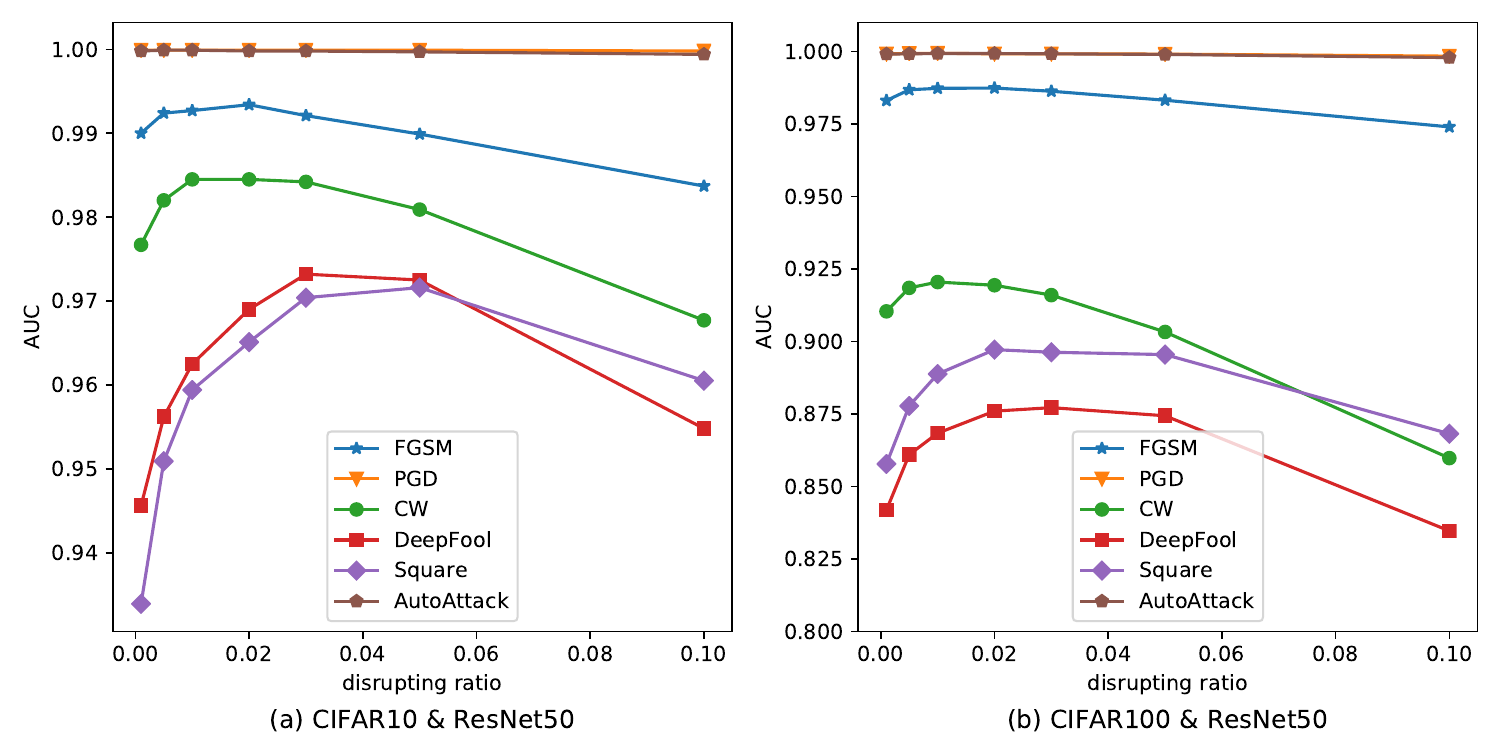}}
  \vspace{-10pt}
  \caption{Sensitivity study. The x label denotes the disruption rate, and y label denotes the corresponding AUC}
  \label{fig7}
  \vspace{-5pt}
\end{figure}

As shown in Fig.\ref{fig7}, with the increase of the disrupting ratio, AUC exhibits an initial rise and then a decreasing trend for FGSM, CW, DeepFool, and Square, while it shows a tiny changes in PGD and AutoAttack. 
Considering the performance of our dynamic features across these six attacks, we choose the disrupting ratio $r=3\%$ in our system.

\section{Conclusion}
To tackle the generalization issues in existing methods, we construct a dynamically stable system from the perspective of Lyapunov stable theory.
By continuously introducing artificial disruption, we extract the inputs stability features to accomplish the detection task. 
We compare our method with seven different detection approaches, and across various model structures and datasets, our method demonstrates optimal detection performance, up to 99.83\%, 97.81\% and 94.47\% average AUC across MNIST, CIFAR10, and CIFAR100. 
Additionally, intensity, generalization, ablation, and sensitivity studies have been conducted, demonstrating the SOTA performance of our method.


\clearpage
\bibliographystyle{named}
\bibliography{ijcai24}

\end{document}